# Active contours driven by local and global intensity fitting energy with application to SAR image segmentation and its fast solvers

Guangming liu
Yantai Science and Technology Association,email:1733823365@qq.com.

**Abstract**─In this paper, we propose a novel variational active contour model based on Aubert-Aujol (AA) denoising model, which hybrides geodesic active contour (GAC) model with active contours without edges (ACWE) model and can be used to segment images corrupted by multiplicative gamma noise. We transform the proposed model into classic ROF model by adding a proximity term. [26] was submitted on 29-Aug-2013, and our early edition was ever submitted to TGRS on 12-Jun-2012, Venkatakrishnan et al. [27] proposed their **"PnP algorithm"** on 29-May-2013, so Venkatakrishnan and we proposed the **"PnP algorithm"** almost simultaneously. Inspired by a fast denosing algorithm proposed by Jia-Zhao recently, we propose two fast fixed point algorithms to solve SAR image segmentation question. Experimental results for real SAR images show that the proposed image segmentation model can efficiently stop the contours at weak or blurred edges, and can automatically detect the exterior and interior boundaries of images with multiplicative gamma noise. The proposed fast fixed point algorithms are robustness to initialization contour, and can further reduce about 15% of the time needed for algorithm proposed by Goldstein-Osher.

## Ⅰ.INTRODUCTION

Image segmentation technique in airborne or satellite-borne SAR images is a powerful tool in environmental monitoring applications such as coastline extraction , object detection and so forth. Recently, level set methods (LSM) have been extensively applied to water/land segmentation in SAR images [1]-[4]. We know that level set methods can be divided into two major categories: region-based models [1-8] and edge-based models [9-12]. In order to make use of edge information and region characteristic, region-based model is often integrated with edge-based model to form the energy functional of models. There are some advantages of level set methods over classical image segmentation methods, such as edge detection, thresholding, and region grow: level set methods can give sub-pixel accuracy of object boundaries, and can be easily formulated under a energy functional minimization, and can provide smooth and closed contour as a result of segmentation and so forth. However, a common difficulty with above variational image segmentation models [1-12] is that the energy functionals to be minimized are all not convex. Thus we may obtain a local minimizer rather than a unique global minimizer from these non-convex energy functionals. This is a more serious problem because local minima of image segmentation often has completely wrong levels of detail and scale. In order to resolve the problems associated with non-convex models, Chan-Esedoglu-Nikolova [13] proposed a convex relaxation approach for image segmentation models. But the globally convex segmentation models contain a total variation (TV) term that is not differentiable in zero, making them difficult to compute. It is well known that the split Bregman method is a technique for fast minimization of $l^1$ regularized energy functional. Goldstein-Osher (GO) [14-15] proposed a more efficient way to compute the TV term by applying the split Bregman method. Recently, Jia-Zhao[16-17] give a algorithm for the solution of ROF denoising model [18] based on anisotropic TV term, which is much faster than GO algorithm.

On the one hand, we know SAR images are affected by speckle, a multiplicative gamma noise that gives the images a grainy appearance and makes the interpretation of SAR images a more difficult task to achieve. Recently, Aubert-Aujol (AA) [19] proposed a non-convex energy functional to remove multiplicative gamma noise based on the maximum a posteriori (MAP) regularization approach. Inspired by [5], we propose a novel variational active contour model based on AA model for SAR image segmentation. On the other hand, it is well known that the ROF denoising model is strictly convex, which always admits a unique solution. By establishing the equivalence relationship between the energy functional of image segmenatation and the weighted ROF denoising model, we further give two fast fixed point algorithms based the algorithm proposed by Jia-Zhao which do not involve partial differential equations.

This paper is organized as follows. Section Ⅱ describes the proposed variational active contour model. Section Ⅲ describes the proposed two fast globally convex segmentation algorithms, and Section Ⅳ describes the experimental results. Section Ⅴ concludes this paper.

## Ⅱ. A NOVEL VARIATIONAL ATIVE CONTOUR MODEL BASED ON AA MODEL

We denote by $R^2$ the usual 2-dimensional Euclidean space $H$. We use $\langle .,. \rangle$ and $\| \, \|$, respectively, to denote the inner product and the corresponding $l^2$ norm of an Euclidean space $H$ while $\| \, \|_1$ is used to denote the $l^1$ norm. We denote by $\nabla_x^T$ ($\nabla_y^T$) the conjugate of the gradient operator $\nabla_x$ ($\nabla_y$).

Given a observed intensity image $f : \Omega \to R (f > 0)$, where $\Omega$ is a bounded open subset of $R^2$, speckle is well modeled as a multiplicative random noise $n$, which is independent of the true image $u$, i.e. $f = u \cdot n$. We know that fully developed multiplicative speckle noise is Gamma distributed with mean value $\mu_n = 1$ and variance $\sigma_n^2 = 1/L$, where $L$ is the equivalent number of independent looks of the image.

**A. Ayed's model**

Ayed et al. [19] proposed a multiregion SAR image segmentation model, which based on level set method and active contour model. The energy functional of the model can be written in the level set formulation as:

$$E(\phi, C_1, C_2) = \int_\Omega \delta(\phi) \left|\nabla\phi\right| dx + \mu \sum_{i=1}^{2} \int_\Omega (\log C_i + \frac{f}{C_i}) M_i(\phi) dx \tag{1}$$

Where $\phi$ is a level set function, $M_1 = H(\phi)$, $M_2 = 1 - H(\phi)$, $H(\phi)$ is the Heaviside function. Then using the level set formulation, the true image $u$ can be expressed：

$$u = C_1 H(\phi) + C_2 (1 - H(\phi)). \tag{2}$$

$H(\phi)$ is often approximated by a smooth function $H_\varepsilon(\phi)$ defined by

$$H_\varepsilon(\phi) = \frac{1}{2}[1 + \frac{2}{\pi}\arctan(\frac{\phi}{\varepsilon})] \tag{3}$$

to automatically detect interior contours and insure the computation of a global minimizer in this paper.

The derivative of $H_\varepsilon(\phi)$ is also approximated by a smooth function

$$\delta_\varepsilon(\phi) = \frac{\partial H_\varepsilon(\phi)}{\partial\phi} = \frac{1}{\pi}(\frac{\varepsilon}{\varepsilon^2 + \phi^2}) \tag{4}$$

Therefore, the energy functional (1) becomes:

$$E_\varepsilon(\phi, C_1, C_2) = \int_\Omega \delta_\varepsilon(\phi) \left|\nabla\phi\right| dx + \mu \sum_{i=1}^{2} \int_\Omega (\log C_i + \frac{f}{C_i}) M_i^\varepsilon(\phi) dx \tag{5}$$

We further modify (5) to incorporate information from an edge detector $g$, which can make (5) more likely to favor segmentation along curves where the edge detector function $g$ is minima. Thus we give a novel hybrid model between region-based method and edge-based method, which can be used to segment image with intensity homogeneity, denoted by GAA, i.e.

$$E_\varepsilon(\phi, C_1, C_2) = \int_\Omega g\delta_\varepsilon(\phi)\left|\nabla\phi\right|dx + \mu\sum_{i=1}^{2}\int_\Omega(\log C_i + \frac{f}{C_i})M_\varepsilon(\phi)dx \tag{6}$$

**B.Proposed model**

According to (6), we further give a local intensity fitting (LIF) energy $E_\varepsilon^{LAA}$ based on region-scalable-fitting (RSF) model[7], which can be used to segment image with intensity inhomogeneity, denoted by LAA, i.e.

$$E_\varepsilon^{LAA}(\phi, C_1, C_2) = \mu\sum_{i=1}^{2}\int_\Omega(\int_\Omega K_\sigma(x-y)(\log f_i(x) + \frac{f(y)}{f_i(x)})M_i^\varepsilon(\phi(y))dy)dx \tag{7}$$

Where $M_1^\varepsilon = H_\varepsilon(\phi)$ , $M_2^\varepsilon = 1 - H_\varepsilon(\phi)$ . $K_\sigma$ is a Gaussian kernel with standard deviation $\sigma$ .

For fixed level set function $\phi$ , we minimize the function $E_\varepsilon^{GAA}(\phi, C_1, C_2)$ with respect to the constant $C_1$ and $C_2$ . By calculus of variations, it is easy to solve them by

$$C_1 = \frac{\int_\Omega f(x)H_\varepsilon(\phi(x))dx}{\int_\Omega H_\varepsilon(\phi(x))dx}, \quad C_2 = \frac{\int_\Omega f(x)(1-H_\varepsilon(\phi(x)))dx}{\int_\Omega(1-H_\varepsilon(\phi(x)))dx}. \tag{8}$$

For fixed level set function $\phi$ , we minimize the function $E_\varepsilon^{LAA}(\phi, f_1, f_2)$ with respect to the constant $f_1(x)$ and $f_2(x)$ . By calculus of variations, it is easy to solve them by

$$f_1(x) = \frac{K_\sigma(x) * [H_\varepsilon(\phi(x))I(x)]}{K_\sigma(x) * H_\varepsilon(\phi(x))}, \quad f_2(x) = \frac{K_\sigma(x) * [(1-H_\varepsilon(\phi(x)))I(x)]}{K_\sigma(x) * [1-H_\varepsilon(\phi(x))]} \tag{9}$$

In the following, we propose a novel hybrid model between GAA model and LAA

model, which can be used to segment image with real SAR images, denoted by GLAA, i.e.

$$E_{\varepsilon}^{GLAA}(\phi, C_1, C_2, f_1, f_2) = \omega E_{\varepsilon}^{GAA}(\phi, C_1, C_2) + (1-\omega) E_{\varepsilon}^{LAA}(\phi, f_1, f_2) \quad (10)$$

Minimization of the energy functional $E_{\varepsilon}^{GLAA}(\phi, C_1, C_2, f_1, f_2)$ with respect to $\phi$ , can be obtained by solving gradient descent flow equation:

$$\frac{\partial \phi}{\partial t} = \delta_{\varepsilon}(\phi) div(g \frac{\nabla \phi}{|\nabla \phi|}) - \mu \cdot \delta_{\varepsilon}(\phi) \cdot (\eta_1 + \eta_2) \quad (11)$$

Where $\eta_1 = \omega[(\log C_1 + \frac{f(x)}{C_1})\text{-}(\log C_2 + \frac{f(x)}{C_1})]$ ,

$$\eta_2 = (1-\omega)[\int_{\Omega} K_{\sigma}(x-y)(\log f_1(y) + \frac{f(x)}{f_1(y)})dy - \int_{\Omega} K_{\sigma}(x-y)(\log f_2(y) + \frac{f(x)}{f_2(y)})dy] .$$

**We denote the level set method for GLAA model as model 1** in this paper.

## Ⅲ. FAST GLOBALLY CONVEX SEGMENTATION MODELS BASED ON THE GLAA MODEL

Based on the globally convex segmentation (GCS) [13] and its application in [14], we can get a simplified flow equation which has the coincident stationary solution with (11) as follows:

$$\frac{\partial \phi}{\partial t} = div(\frac{\nabla \phi}{|\nabla \phi|}) - \mu \cdot (\eta_1 + \eta_2) \quad (12)$$

The simplied flow represent gradient descent for minimizing the energy functional:

$$F^{GLAA}(\phi) = \|\nabla \phi\|_1 + \mu < \phi, \eta > \quad (13)$$

Where $\|\nabla \phi\|_1 = \int_{\Omega} |\nabla \phi| dx$ , $< \phi, \eta >= \int_{\Omega} \phi \cdot \eta dx$ , $\eta = \eta_1 + \eta_2$ .

In order to guarantee the unique global minimizer of energy functional (13), we restrict the solution $\phi$ to lie in a finite interval ( $0 \le \phi \le 1$ ).We suppose the weighted TV term of the energy functionals (13) is anisotropic. By applying the anisotropic TV term to (13), we can obtain a globally convex segmentation model as follows:

$$F^{GCGLAA}(\phi) = \min_{0 \le \phi \le 1} \|\nabla_x \phi\|_1 + \|\nabla_y \phi\|_1 + \mu < \phi, \eta > \tag{14}$$

**A. Split Bregman method for GCGLAA model**

In the past, solutions of the TV model were based on nonlinear partial differential equations and the resulting algorithms were very complicated. It is well known that the split Bregman method is a technique for fast minimization of $l^1$ regularized energy functional. A break through was made by Goldstein-Osherin (GO) [14-15].They proposed a more efficient way to compute the TV term based on the split Bregman method. To apply the split Bregman method to (14), we introduce auxiliary variables $d_x \leftarrow \nabla_x \phi$, $d_y \leftarrow \nabla_y \phi$, and add a quadratic penalty function to weakly enforce the resulting equality constraint which results in the following unconstrained problem:

$$\begin{aligned} (\phi^{k+1}, d_x^{k+1}, d_y^{k+1}) &= \arg\min_{0 \le \phi \le 1} \|d_x\|_1 + \|d_y\|_1 + \mu < \phi, \eta > + \\ & \frac{\lambda}{2}\|d_x - \nabla_x \phi\|^2 + \frac{\lambda}{2}\|d_y - \nabla_y \phi\|^2 \end{aligned} \tag{15}$$

We then apply split Bregman method to strictly enforce the constraints $d_x = \nabla_x \phi$, $d_y = \nabla_y \phi$. The resulting optimization problem becomes:

$$\begin{aligned} (\phi^{k+1}, d_x^{k+1}, d_y^{k+1}) &= \arg\min_{0 \le \phi \le 1} \|d_x\|_1 + \|d_y\|_1 + \mu < \phi, \eta > + \\ & \frac{\lambda}{2}\|d_x - \nabla_x \phi - b_x^k\|_2^2 + \frac{\lambda}{2}\|d_y - \nabla_y \phi - b_y^k\|_2^2 \\ b_x^{k+1} &= b_x^k + \nabla_x \phi^{k+1} - d_x^{k+1} \\ b_y^{k+1} &= b_x^k + \nabla_y \phi^{k+1} - d_y^{k+1} \end{aligned} \tag{16}$$

For fixed $\vec{d}$, the Euler-Lagrange equation of optimization problem (16) with respect to $\phi$ is:

$$\Delta \phi^{k+1} = \frac{\mu \eta}{\lambda} - \nabla_x^T (d_x^k - b_x^k) - \nabla_y^T (d_y^k - b_y^k) \tag{17}$$

For fixed $\phi$, minimization of (16) with respect to $\vec{d}$ gives:

$$
\begin{aligned}
d_x^{k+1} &= shrink_{1/\lambda}(\nabla_x \phi^{k+1} + b_x^k) \\
d_y^{k+1} &= shrink_{1/\lambda}(\nabla_x \phi^{k+1} + b_y^k)
\end{aligned} \tag{18}
$$

Where $shink_{1/\lambda}(x) = \mathrm{sgn}(x)\max(|x| - 1/\lambda, 0)$ .

By using central discretization for Laplace operator and backward difference for divergence operator, the numerical scheme for (18) becomes:

$$
\begin{aligned}
\alpha_{i,j} &= d^x_{i-1,j} - d^x_{i,j} - b^x_{i-1,j} + b^x_{i,j} + d^y_{i,j-1} - d^y_{i,j} - b^y_{i,j-1} + b^y_{i,j} \\
\beta_{i,j} &= \frac{1}{4}(\phi_{i-1,j} + \phi_{i+1,j} + \phi_{i,j-1} + \phi_{i,j+1} - \frac{\mu \cdot \eta}{\lambda} + \alpha_{i,j}) \\
\phi_{i,j} &= \max(\min(\beta_{i,j}, 1), 0)
\end{aligned} \tag{19}
$$

As the optimal $\phi$ is found, the segmented region can be found by thresholding the level set function $\phi(x)$ for some $\gamma \in (0,1)$ : $\Omega_1 = \{x : \phi(x) > \gamma\}$ .

The split Bregman algorithm for the minimization problem (16) can be summarized as follows(**model 2**):

Given: noisy image $f$ ; $\lambda > 0$ , $\mu > 0$

Initialization: $b^0 = 0$ , $d^0 = 0$ , $\phi^0 = f / \max(f)$

For $k = 0,1,2,\cdots$

compute (19)

$\phi^{k+1} = \max(\min(\phi^k, 1), 0)$

$d_x^{k+1} = shrink_{1/\lambda}(\nabla_x \phi^{k+1} + b_x^k)$

$d_y^{k+1} = shrink_{1/\lambda}(\nabla_x \phi^{k+1} + b_y^k)$

$b_x^{k+1} = b_x^k + \nabla_x \phi^{k+1} - d_x^{k+1}$

$b_y^{k+1} = b_x^k + \nabla_y \phi^{k+1} - d_y^{k+1}$

END

B. Fixed point algorithm 1 for GCGLAA model

We know the energy functional of image segmentation (14) does not have a unique global minimizer because it is homogeneous degree one. However, the ROF denoising model always admits a unique solution because the energy functional is strictly convex. In order to utilizing the convexity of ROF denoising model and favor segmentation along curves where the edge detector function is minima, we give a discrete version weighted ROF (WROF) denoising model based on anisotropic TV term as follows:

$$E_{WROF}(\phi)=\left\|\nabla_x\phi\right\|_1+\left\|\nabla_y\phi\right\|_1+\frac{\alpha}{2}\left\|\phi-f\right\|^2 \tag{20}$$

Where $\alpha>0$ is an appropriately chosen positive parameter.

Note that GO algorithm [14-15] still requires solving a partial difference equation in each iteration step. Recently, Jia-Zhao (JZ)[16-17] proposed a fast algorithm for image denoising based on TV term, which is very simple and does not involve partial differential equations or difference equation. We further transform the proposed model into a generalized ROF model by adding a proximity term ,and it can be solved by a fast denoising algorithm proposed by Jia-Zhao or soved by BM3D and NLM denoising algorithm, which also provided a unified solution framework for formally generalized-ROF-like subproblems arising in multivariate splitting algorithms.

In order to apply the JZ algorithm to image segmentation question (14), we first suppose that $\phi^k$ and $\eta^k$ are known and reformulate (14) by adding a proximity term $\frac{\alpha}{2}\left\|\phi-\phi^k\right\|^2$ as:

$$\begin{aligned}\phi^{k+1}&=\arg\min_{0\le\phi\le1}\left\|\nabla_x\phi\right\|_1+\left\|\nabla_y\phi\right\|_1+\mu<\phi,\eta>+\frac{\alpha}{2}\left\|\phi-\phi^k\right\|^2\\&=\arg\min_{0\le\phi\le1}\left\|\nabla_x\phi\right\|_1+\left\|\nabla_y\phi\right\|_1+\mu<\phi-\phi^k,\eta>+\frac{\alpha}{2}\left\|\phi-\phi^k\right\|^2\\&=\arg\min_{0\le\phi\le1}\left\|\nabla_x\phi\right\|_1+\left\|\nabla_y\phi\right\|_1+\frac{\alpha}{2}\left\|\phi-(\phi^k-\frac{\mu\eta}{\alpha})\right\|^2\end{aligned} \tag{21}$$

Treating $\phi^k-\frac{\mu\eta}{\alpha}$ as the **"noisy image"**, (21) minimizes the residue between the **"clean image"** $\phi$ and $\phi^k-\frac{\mu\eta}{\alpha}$ using the total variation norm, then (21) is the total variation denoising problem. So we propose a fixed point algorithm 1 **(model 3)** based on JZ algorithm to solve (21) as follows:

$$
\begin{aligned}
b_x^k &= (I - shrink_{1/\lambda})(\nabla_x \phi^k + b_x^{k-1}) \\
b_y^k &= (I - shrink_{1/\lambda})(\nabla_y \phi^k + b_y^{k-1}) \\
\phi^{k+1} &= \phi^k - \frac{\mu\eta}{\alpha} - \frac{\lambda}{\alpha}(\nabla_x^T b_x^k + \nabla_y^T b_y^k) \\
\phi^{k+1} &= \max(\min(\phi^{k+1}, 1), 0)
\end{aligned}
\tag{22}
$$

According to [22-23], we infer from (22) that operator $(I - shrink_{1/\lambda})(I - \frac{\lambda}{\alpha}\nabla\nabla^T)$ is nonexpansive when $\frac{\lambda}{\alpha}$ is less than $\frac{1}{4}\sin^{-2}\frac{(N-1)\pi}{2N}$ which is slightly bigger than 1/4 . In order to accelerate the convergence of (23) , we adopt the following iteration scheme by utilizing k-averaged operator theory ( see more details in [23]):

$$
\begin{aligned}
b_x^k &= t \cdot b_x^{k-1} + (1-t)\cdot(I - shrink_{1/\lambda})(\nabla_x \phi^k + b_x^{k-1}) \\
b_y^k &= t \cdot b_y^{k-1} + (1-t)\cdot(I - shrink_{1/\lambda})(\nabla_y \phi^k + b_y^{k-1})
\end{aligned}
\tag{23}
$$

Where the weight factor $t \in (0,1)$ is called the relaxation parameter. As the optimal $\phi(x)$ is found, the segmented region can be found by thresholding the function $\phi(x)$ for some $0 < \gamma < 1$: $\Omega_1 = \{x : \phi(x) > \gamma\}$ . The FPA 1 for the minimization problem (21) can be summarized as follows:

Given: noisy image $f$ ; $\lambda > 0$ , $\mu > 0$ , $\alpha > 0$ , $t \in (0,1)$ , $\gamma \in (0,1)$

Initialization: $b_x^0 = 0$ , $b_y^0 = 0$ , $\phi^0 = f / \max(f)$ , $\Omega_0 = \{x : \phi^0 > \gamma\}$ ,he

$C_1^0 = \int_{\Omega_0} f dx$ , $C_2^0 = \int_{\Omega_0^c} f dx$

For $k = 0,1,2,\cdots$

$$
\begin{aligned}
b_x^k &= t \cdot b_x^{k-1} + (1-t)\cdot(I - shrink_{1/\lambda})(\nabla_x \phi^k + b_x^{k-1}) \\
b_y^k &= t \cdot b_y^{k-1} + (1-t)\cdot(I - shrink_{1/\lambda})(\nabla_y \phi^k + b_y^{k-1}) \\
\phi^{k+1} &= \phi^k - \frac{\mu\eta}{\alpha} - \frac{\lambda}{\alpha}(\nabla_x^T b_x^k + \nabla_y^T b_y^k) \\
\phi^{k+1} &= \max(\min(\phi^{k+1}, 1), 0) \\
\Omega_{k+1} &= \{x : \phi(x) > \gamma\}, \quad C_1^{k+1} = \int_{\Omega_{k+1}} f dx, \; C_2^{k+1} = \int_{\Omega_{k+1}^c} f dx
\end{aligned}
$$

END

C. Fixed point algorithm 2 for GCGLAA model

We can also reformulate (14) by introducing a term $\frac{\alpha}{2}\|\phi-\varphi\|^2$ as:

$$(\varphi^{k+1},\phi^{k+1}) = \arg\min_{\varphi,0\le\phi\le 1} \|\nabla_x\phi\|_1 + \|\nabla_y\phi\|_1 + \mu < \phi,\eta > + \frac{\alpha}{2}\|\phi-\varphi\|^2 \qquad (24)$$

We then apply split Bregman method to (24) and propose a fixed point algorithm 2 (**model 4**) as follows:

$$\begin{aligned}
\varphi^k &= \arg\min_{0\le\varphi\le 1} \frac{\alpha}{2}\|\phi^k-\varphi-c^{k-1}\|^2 + \mu < \varphi,\eta > \\
c^k &= c^{k-1} + \varphi^k - \phi^k \\
\phi^{k+1} &= \arg\min_{0\le\phi\le 1} \|\nabla_x\phi\|_1 + \|\nabla_y\phi\|_1 + \frac{\alpha}{2}\|\phi-\varphi^k-c^k\|^2
\end{aligned} \qquad (25)$$

we can get a solution of (25) as **model 3**,i.e.

$$\begin{aligned}
\varphi^k &= \phi^k - c^{k-1} - \frac{\mu}{\alpha}\eta, \\
\varphi^k &= \max(\min(\varphi^k,1),0) \\
c^k &= c^{k-1} + \varphi^k - \phi^k \\
\phi^{k+1} &= \varphi^k + c^k - \frac{\lambda}{\alpha}(\nabla_x^T b_x^k + \nabla_y^T b_y^k)
\end{aligned} \qquad (26)$$

Where $b_x^k$ and $b_y^k$ is also defined as (23). As the optimal $\varphi$ is found, the segmented region can be found by thresholding the function $\varphi(x)$ for some $\gamma\in(0,1)$: $\Omega_1=\{x:\varphi(x)>\gamma\}$. The FPA 2 (**model 4**)for the minimization problem (25) can be summarized as follows:

Given: noisy image $f$; $\lambda>0$, $\mu>0$, $\alpha>0$, $t\in(0,1)$, $\gamma\in(0,1)$

Initialization: $b_x^0=0$, $b_y^0=0$, $c^0=0$, $\phi^0=f/\max(f)$

$\Omega_0=\{x:\phi^0>\gamma\}$, $C_1^0=\int_{\Omega_0} f dx$, $C_2^0=\int_{\Omega_0^c} f dx$

For $k=0,1,2,\cdots$

$b_x^k = t\cdot b_x^{k-1} + (1-t)\cdot(I-shrink_{g/\lambda})(\nabla_x\phi^k + b_x^{k-1})$

$b_y^k = t\cdot b_y^{k-1} + (1-t)\cdot(I-shrink_{g/\lambda})(\nabla_y t^k + b_y^{k-1})$

$$\varphi^k = \phi^k - c^{k-1} - \frac{\mu}{\alpha}\eta,$$

$$\varphi^k = \max(\min(\varphi^k, 1), 0)$$

$$c^k = c^{k-1} + \varphi^k - \phi^k$$

$$\phi^{k+1} = \varphi^k + c^k - \frac{\lambda}{\alpha}(\nabla_x^T b_x^k + \nabla_y^T b_y^k)$$

$$\Omega_{k+1} = \{x : \varphi(x) > \gamma\},\ C_1^{k+1} = \int_{\Omega_{k+1}} f dx,\ C_2^{k+1} = \int_{\Omega_{k+1}^c} f dx$$

END

**D. Quantitative Evaluation:**

We give the following several measures to evaluate the proposed FPA algorithm in this paper .

**1) Uniformity measurement**

We adopt the uniformity measurement of image segmentation regions to evaluate the performance of the proposed method. The interior of each region should be uniform after the segmentation and there should be a great difference among different regions. That is to say, the uniformity degree of regions represents the quality of the segmentation. Therefore, we give the measurement of segmentation accuracy (SA) as follows[25]:

$$pp = 1 - \frac{1}{C}\sum_i \{\sum_{x\in R_i}[f(x) - \frac{1}{A_i}\sum_{x\in R_i} f(x)]^2\} \quad (27)$$

Where $R_i$ denotes different segmentation regions, $C$ is the normalization constant, $f(x)$ is the gray value of point $x$ in the image, $A_i$ is the number of the pixels in each region $R_i$. The closer to 1 the value of $pp$ is, the more uniform the interior of the segmentation regions are and the better the quality of the segmentation is.

**2). Dice similarity coefficient**

The ground truth (GT) is drawn manually through visual inspection of the SAR images. The Dice Similarity coefficient (DSC) [24] between the computed segmentation (CS) and the GT is defined as:

$$DSC(CS,GT) = 2 \times \frac{N(CS \cap GT)}{N(CS) + N(GT)} \tag{28}$$

Where $N(\cdot)$ indicates the number of voxels in the enclosed set. The closer the DSC value to 1, the better the segmentation.

## Ⅳ. EXPERIMENTAL RESULTS

We test our model with synthetic and real images in this section. Two synthetic images (by setting $L = 2$ for Gamma noise) are used to test the efficiency of our proposed model, whose size are $85 \times 76$ and $85 \times 61$, respectively. Two ERS-2 SAR images have the size $398 \times 344$ and $240 \times 279$, respectively, and the gray-scale in the range between 0 and 255.

The level set function can be simply initialized as a binary step function which takes a constant value 1 inside a region and another constant value -1 outside. We compare our proposed **model1-model4** with model in [4], the model in [5] and the globally convex segmentation model in [14]. All the models are implemented with Matlab 8.0 in core2 with 1.9 GHZ and 1GB RAM.

For synthetic image1 and image 2, we adopt the following parameters.

The parameters of the model [4] are chosen as $\mu = 1$, $\Delta t = 0.1$, $\varepsilon = 1$, $\sigma = 20$. The parameters of model in [5] are chosen as $\mu = 0.04$, $\Delta t = 0.1$, $\varepsilon = 1$. The parameters of the model [14] are chosen as $\mu = 1$, $\Delta t = 0.1$, $\varepsilon = 1$, $\sigma = 20$ .The parameters of proposed model 1 are chosen as $\mu = 255$, $\Delta t = 0.1$, $\varepsilon = 1$ .The parameters of proposed model 2 are chosen as $\mu = 20$ $\lambda = 0.02$, $\alpha = 0.2$ .The parameters of proposed model 3(4) are chosen as $\mu = 20$, $\lambda = 0.02$, $\alpha = 0.2$, $t = 1e-5$. The thresholding values for proposed model 2-4 are all chosen as $\gamma = 0.5$, which are used to find the segmented region $\Omega_1 = \{x : \phi(x) > \gamma\}$.

For synthetic image 1 and image 2, we compare the speed of algorithms ( the pair (.,.) is used to report both the number of iterations ( the first number ) and the cpu time ( the second number )) and $DSC$, which are listed in TABLE Ⅰ.

We show the final contours of two synthetic images in Fig.1 and Fig.2. We can see from Fig.1 and Fig.2 that, the model in [4] and the model in [14] all can not segment synthetic images correctly, the model in [4] can segment synthetic images correctly, and the proposed model 1 can also automatically detect the exterior and interior boundaries of two synthetic images better than the model [4], We also observe that the proposed model 2 can reduce much time than the proposed model 1, and the proposed model 3 can further reduce about half of the running time needed for the proposed model 1. As can be seen from TABLE Ⅰ, the $DSC$ values of the proposed model 1 are higher than those of the model in [4].

For SAR image1 and image2, we compare the speed of algorithms ( the pair (.,.) is used to report both the number of iterations ( the first number ) and the cpu time ( the second number )) and $pp$, which are listed in TABLE Ⅱ.

We show the final contours of three SAR images in Fig.3 and Fig.4. We can see from Fig.3 and Fig.4 that, the model in [5] and the model in [14] all can not segment synthetic images correctly, the model in [4] can segment synthetic images correctly, and the proposed model 1 can also automatically detect the exterior and interior boundaries of three SAR images with severe multiplicative gamma noise better than the model in [4]. We also observe that the proposed model 2 can reduce much time than the proposed model 1, and the proposed model 3 can further reduce about half of the running time needed for the proposed model 1.

As can be seen from TABLE Ⅱ, the $pp$ values of the proposed model 2, the proposed model 3(4) are all close to 1, which show high precision of the proposed fast models.

# Ⅴ.CONCLUSION

In this paper, we propose a novel variational active contour model based on AA denoising model, which hybrides GAC model with ACWE model and can be used to segment images corrupted by multiplicative gamma noise. We transform the proposed model into classic ROF model by adding a proximity term. Inspired by a fast denoising algorithm proposed by Jia-Zhao recently, we propose two fast fixed point algorithms to solve SAR image segmentation question. Experimental results for real SAR images show that the proposed image segmentation model can efficiently stop the contours at weak or blurred edges, and can automatically detect the exterior and interior boundaries of images with multiplicative gamma noise. The proposed fast fixed point algorithms are robustness to initialization contour, and can further reduce about 15% of the time needed for algorithm proposed by Goldstein-Osher.

The reason why the proposed algorithms are faster than the SGA algorithm are that they do not involve partial differential or difference equations. The proposed algorithms can also be applied to the case of isotropic TV, and all the segmentation question are solved by the ROF model which has been intensively studied.

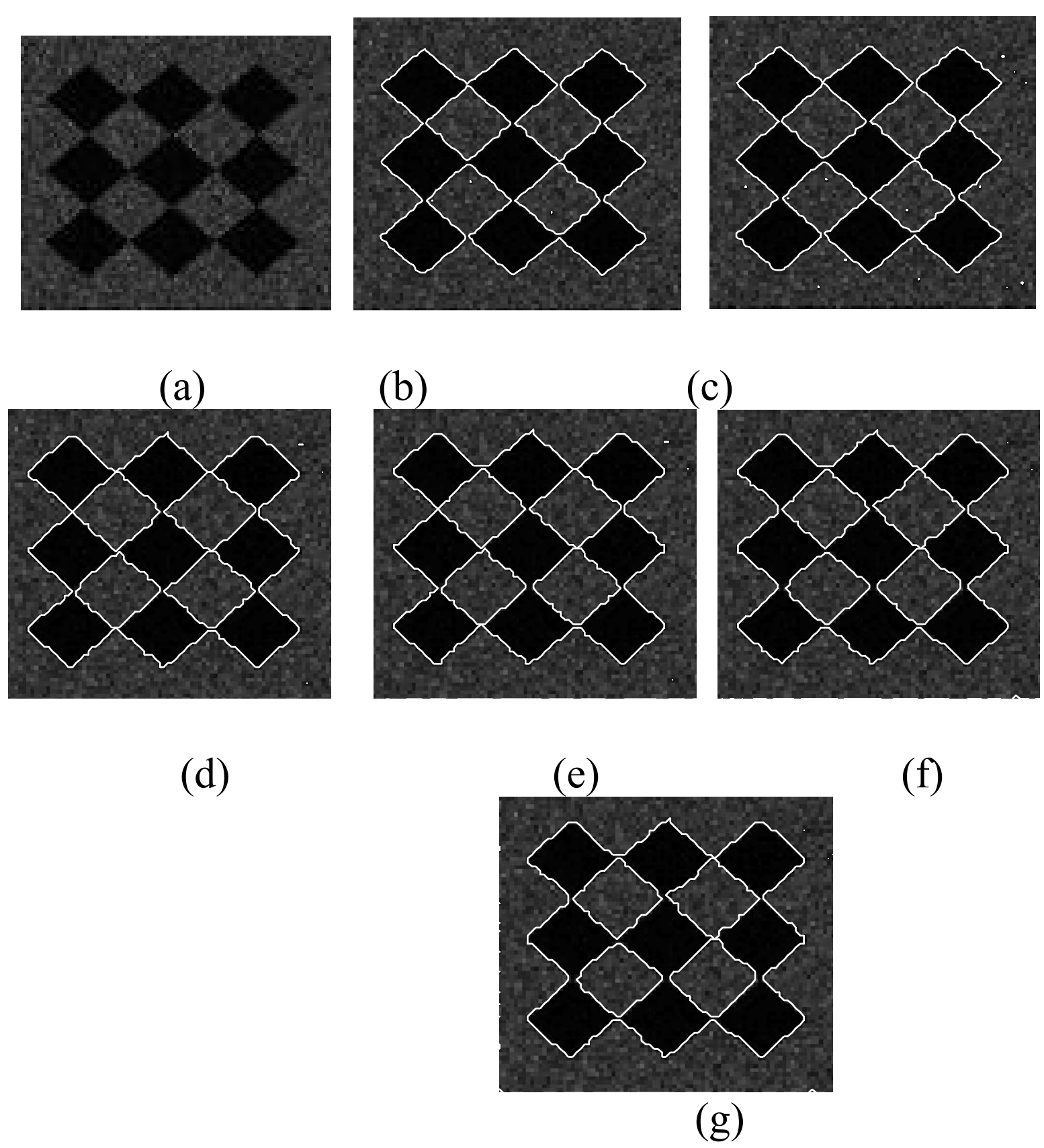

Fig.1. (a) synthetic image 1. (b) final contour by the model in [5].(c) final contour by the model in [14]. (d) final contour by the model in [4] (e) final contour by the proposed model 1.(f) final contour by the proposed model 2.. (g) final contour by the proposed model 3.

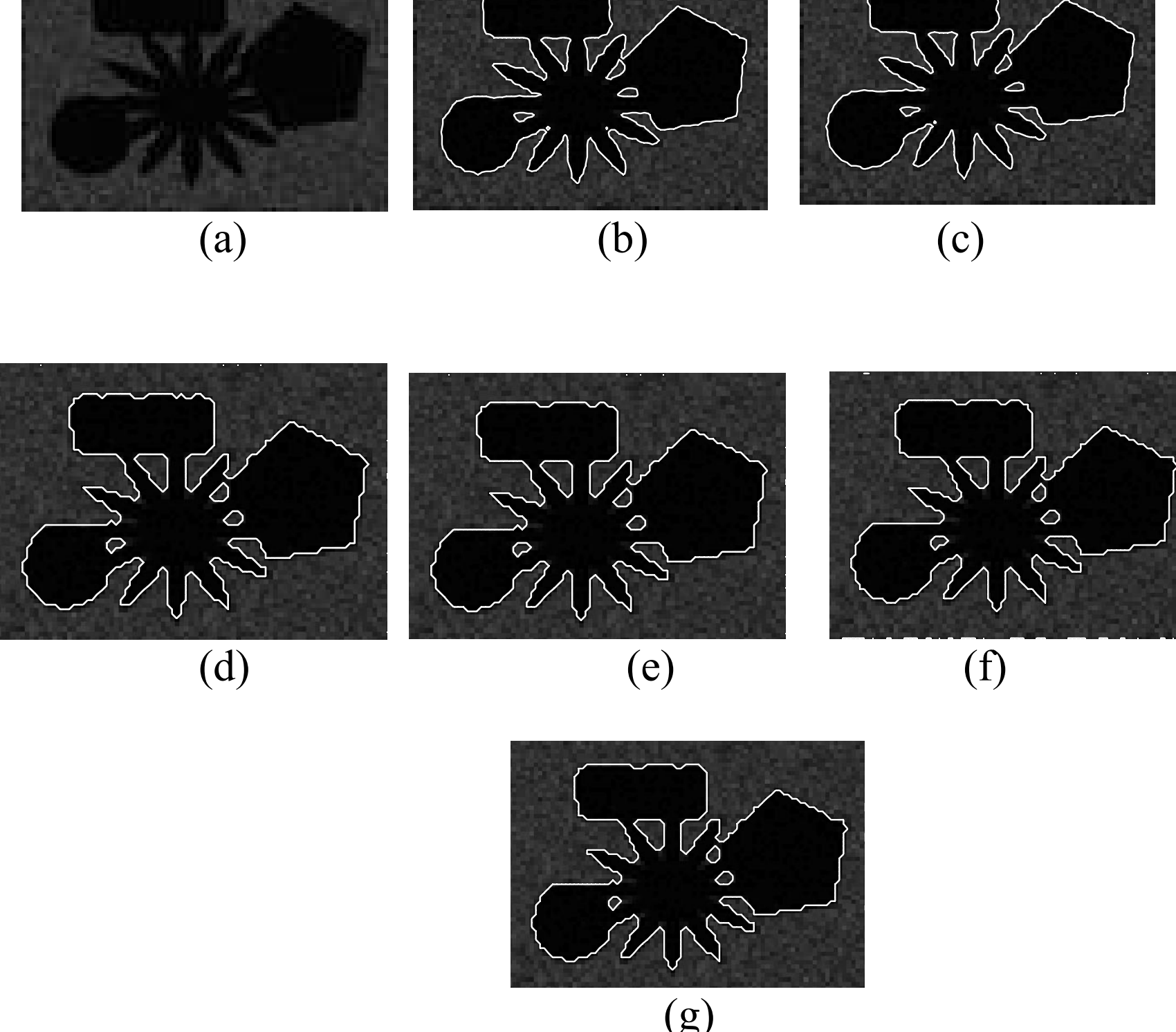

Fig.2. (a) synthetic image 2. (b) final contour by the model in [5].(c) final contour by the model in [14]. (d) final contour by the model in [4] (e) final contour by the proposed model 1.(f) final contour by the proposed model 2.. (g) final contour by the proposed model 3.

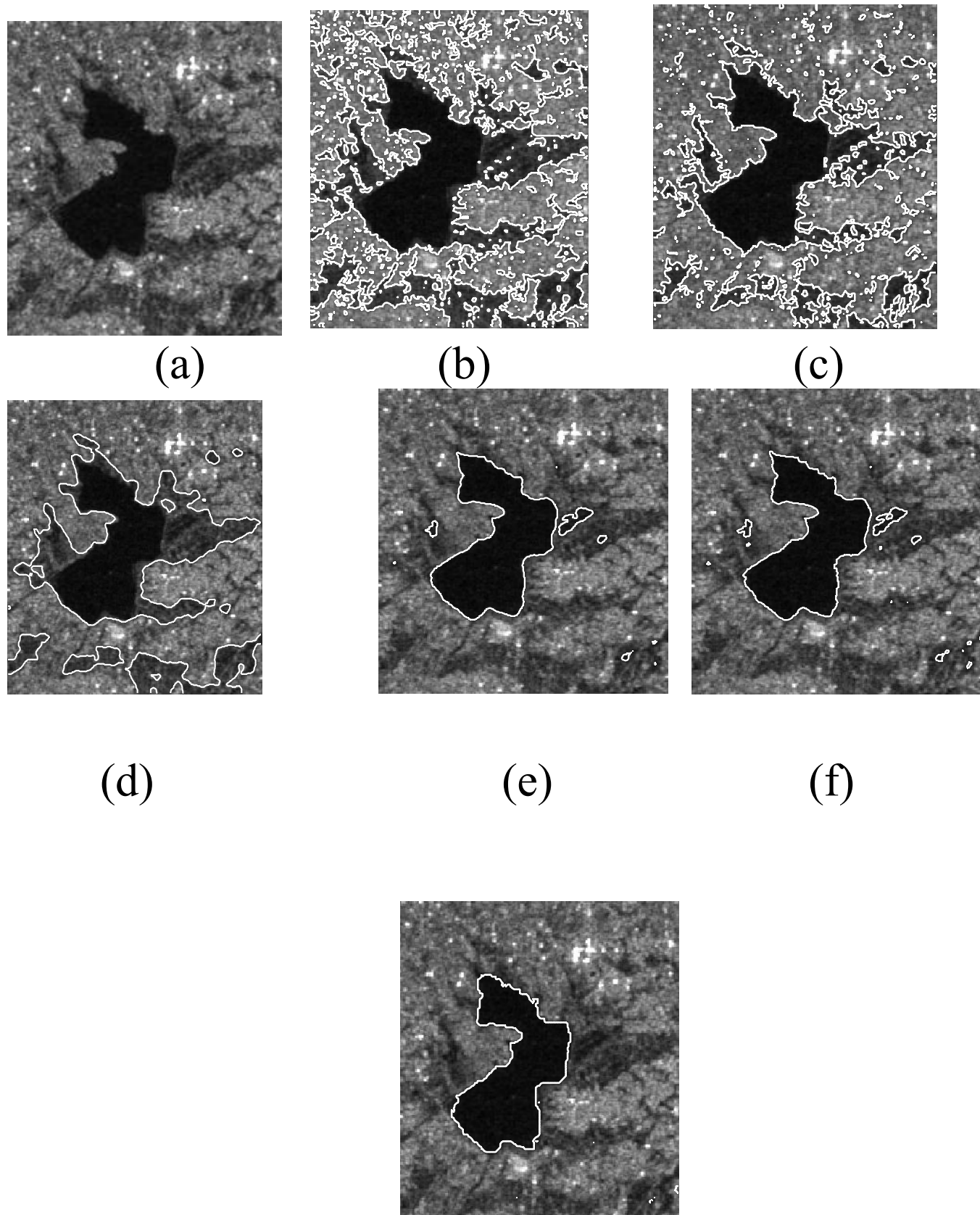

(g)

Fig.3. (a) SAR image1. (b) final contour by the model in [5].(c) final contour by the model in [14]. (d) final contour by the model in [4] (e) final contour by the proposed model 1.(f) final contour by the proposed model 2.. (g) final contour by the proposed model 3.

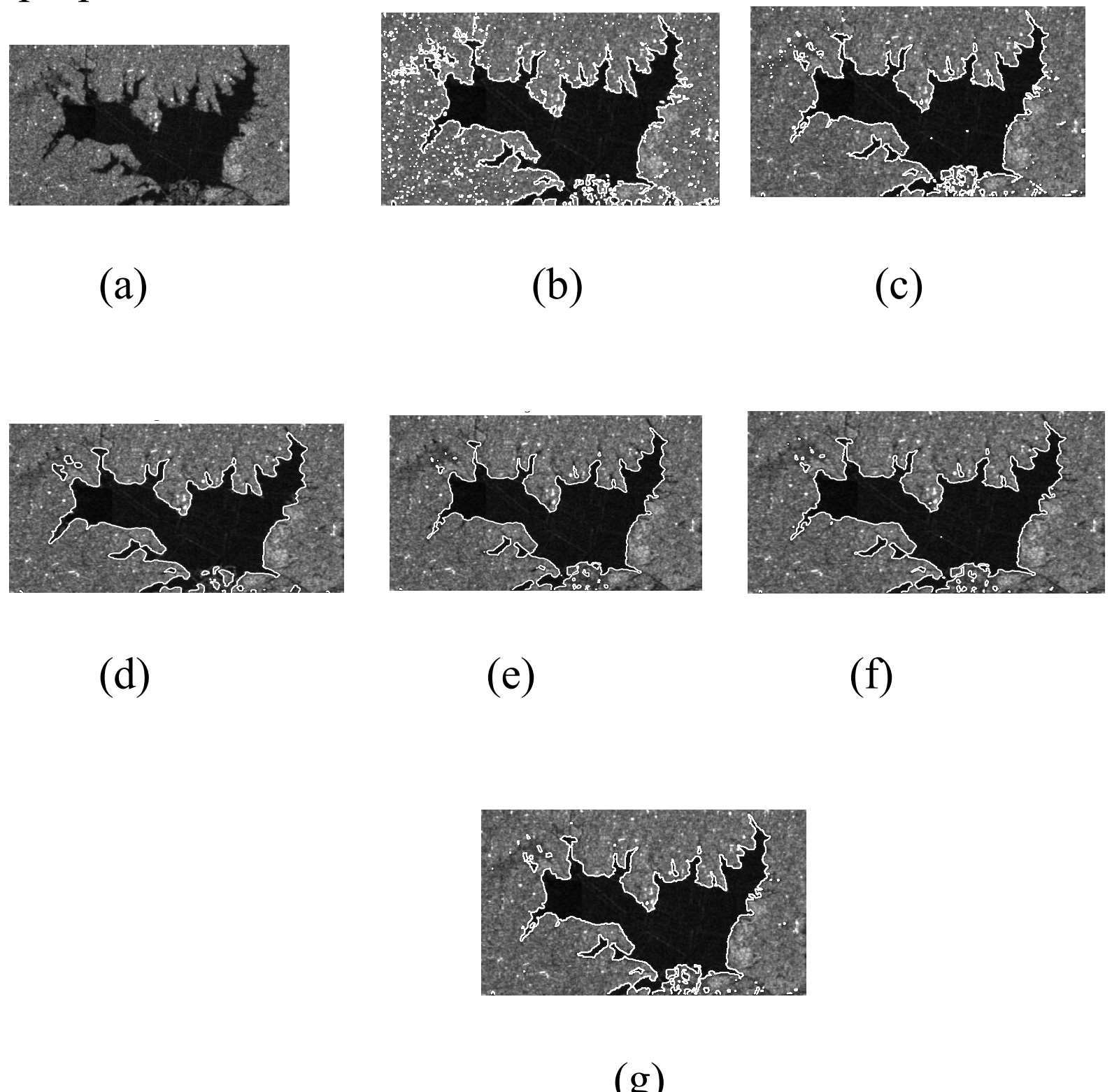

Fig.4. (a) SAR image2. (b) final contour by the model in [5].(c) final contour by the model in [14]. (d) final contour by the model in [4] (e) final contour by the proposed model 1.(f) final contour by the proposed model 2.. (g) final contour by the proposed model 3.

TABLE Ⅰ
PERFORMANCE OF THE PROPOSED MODELS AND CLASSIC MODELS

| Method | synthetic image1 | | synthetic image 2 | |
|---|---|---|---|---|
| | Speed | DSC(PP) | Speed | DSC(PP) |
| Model [5] | (20,0.531s) | 97.37% (0.958) | (20,0.39s) | 97.55% (0.9295) |
| Model [14] | (20,0.422s) | 98.58% (0.932) | (20,0.36s) | 99.11% (0.9133) |
| Model [4] | (10,0.375s) | 96.84% (0.9993) | (10,0.344s) | 97.69% (0.9992) |
| Model 1 | (10,0.359s) | 96.70% (0.9993) | (10,0.344s) | 97.59% (0.9992) |
| Model 2 | (10,0.188s) | 95.88% (0.9993) | (10,0.176s) | 96.90% (0.9992) |

| Model 3 | (10,0.175s) | 95.76% (0.9993) | (10,0.18s) | 96.78% (0.9992) |
|---|---|---|---|---|

TABLE II
PERFORMANCE OF THE PROPOSED MODELS AND CLASSIC MODELS

| Method | SAR image1 | | SAR image 2 | |
|---|---|---|---|---|
| | Speed | PP | Speed | PP |
| Model [5] | (40,5.031s ) | 0.7821 | (40,13.172s) | 0.8268 |
| Model [14] | (50,17.625s ) | 0.9875 | (50,8.734) | 0.9784 |
| Model [4] | (40,5.156s ) | 0.9486 | (40,13.297s ) | 0.9475 |
| Model 1 | ( 30,9.046s ) | 0.999 | (30,4.753s ) | 0.999 |
| Model 2 | (30,4.362s ) | 0.999 | (30,2.439s ) | 0.999 |
| Model 3 | (30,4.681s ) | 0.999 | (30,2.354s ) | 0.999 |